\documentclass{article}

\PassOptionsToPackage{numbers, compress}{natbib}

\usepackage[final]{neurips_2021}

\usepackage[utf8]{inputenc} 
\usepackage[T1]{fontenc}    
\usepackage{hyperref}       
\usepackage{url}     
\usepackage{amsmath}
\usepackage{booktabs}       
\usepackage{amsfonts}       
\usepackage{nicefrac}       
\usepackage{microtype}      
\usepackage{xcolor}         
\usepackage{multirow}
\usepackage{graphicx}
\title{WeightScale: Interpreting Weight Change \\ in Neural Networks}
                    
\author{%
  Ayush Manish Agrawal\thanks{Equal Contributors}\hspace{0.2cm}\footnotemark[2]\\
  University of Nebraska\\
  \texttt{aagrawal@nebraska.edu} \And
  Atharva Tendle\footnotemark[1]\hspace{0.2cm}\thanks{OpenMined $\&$ Manifold Computing}\\
  University of Nebraska-Lincoln\\
  \texttt{atharva.tendle@huskers.unl.edu}\And
  Harshvardhan Sikka\footnotemark[2]\\
  Georgia Institute of Technology\\
  \texttt{harsh@manifold.com}\And
  Sahib Singh\footnotemark[2]\\
  Ford Motor Company\\
  \texttt{sahibsingh570@gmail.com}
}

\begin{document}

\maketitle

\begin{abstract}
Interpreting the learning dynamics of neural networks can provide useful insights into how networks learn and the development of better training and design approaches. We present an approach to interpret learning in neural networks by measuring relative weight change on a per layer basis and dynamically aggregating emerging trends through combination of dimensionality reduction and clustering which allows us to scale to very deep networks. We use this approach to investigate learning in the context of vision tasks across a variety of state-of-the-art networks and provide insights into the learning behavior of these networks, including how task complexity affects layer-wise learning in deeper layers of networks.
\end{abstract}

\section{Introduction}
\label{introduction}
Deep learning based approaches have achieved excellent performance in a variety of problem areas,and generally consist of neural network based models that learn mappings between task specific data and corresponding solutions. The success of these methods relies on their ability to learn multiple representations at different levels of abstraction, achieved through the composition of non-linear modules that transform incoming representations into new ones [1]. These transformation modules are referred to as layers of the neural network, and neural networks with several such layers are referred to as deep neural networks. Significant research has demonstrated the capacity for deep networks to learn increasingly complex functions, often through the use of the specific neural network primitives that introduce information processing biases in the problem domain. For example, in the vision domain, Convolutional Neural Networks (CNNs) utilize convolution operations that use filtering to detect local conjunctions of features in images, which often have local values that are highly correlated and invariant to location in the image.
Alongside this, a general observation in many computer vision tasks is that early layers converge to simple feature configurations \cite{yosinski2014transferable}. This phenomena is observed in many vision architectures, including Inception and Residual Networks \cite{cammarata2020thread:}. These findings, among others, point to a natural question: Do different layers in neural networks converge to their learned features at different times in the training process?

Understanding the layer-wise learning dynamics that allow for a deep neural network to learn the solution of a particular task is of significant interest, as it may provide insight into understanding potential areas of improvement for these algorithms and reduce their overall training costs. 
Our contributions is two-fold:

\begin{itemize} 

    \item We empirically investigate the learning dynamics of different layers in ResNet50 and EfficientNet-B4 across CIFAR-10, CIFAR-100 and SVHN dataset using the Relative Weight Change (RWC) metric \cite{agrawal2021investigating}.

    \item We develop an approach to empirically analyze RWC trends within networks using clustering and dimensionality reduction. This framework allows us to scale RWC analysis to very deep networks and find similarities amongst numerous relative weight change trends something which was infeasible earlier with the original RWC approach \cite{agrawal2021investigating} given the scale and complexity of such networks. 
\end{itemize}

The rest of this text is organized as follows: Section 2 presents related work. Section 3 introduces relative weight change, clustering and our experimental methodology. Section 4 discusses empirical results across several datasets and architectures. Finally, Section 5 discusses conclusions and future directions for this line of research.

\section{Related Work}
\label{related-works}
While Deep Learning explainability is an active area of research, there has been limited research towards understanding the layer level trends in neural networks. Most of the work done so far in this context has been focused towards feature visualization of neural networks \cite{li2020visualizing, Erhan2009VisualizingHF, simonyan2014deep, nguyen2016multifaceted, nguyen2019understanding, zeiler2014visualizing, olah2017feature, Szegedy_2015_CVPR}. While our work is similar in context, our work is concentrated on approaching these features from an empirical standpoint. Instead of focusing on feature visualization we focus on understanding the weights in each layer of the neural network and compute the relative change in these weights across epochs. 
Our work builds upon the earlier proposed work by \cite{agrawal2021investigating} who conceptualized Relative Weight Change (RWC) as a metric to track the change in weight across a given neural network layer for every epoch. They propose RWC as a proxy for layer-wise learning, with the assumption that when the weights of a network have minimal change over a set of epochs, they are converging to their optimum. Our work goes above and beyond by analyzing RWC for much more complex networks than what the paper proposed including ResNet50 and EfficientNet-B4. Realizing the issues which come along with deeper networks due to scale and complexity we also propose a novel solution using K-Means clustering to improve the interpretability aspects of RWC.

\section{Method}

\subsection{RWC}
\label{rwc}
To better understand the layer-wise learning dynamics, we utilize a metric known as Relative Weight Change (RWC) \cite{agrawal2021investigating}. RWC represents the average of the absolute value of the percent change in the magnitude of a given layer's weight. 

\begin{equation} 
\label{RWC}
\centering 
RWC_{L} = \frac{||w_{t} - w_{t-1} ||_{1}}{|| w_{t-1}||_{1}}
\end{equation}
where $L$ represents a single layer in a deep neural network, and $w_{t}$ represents the vector of weights associated with $L$ at a given training step $t$. The $L_{1}$ norm characterizes the difference in magnitude of the weights. This difference is normalized by dividing by the magnitude of the layer's weights during the previous training step. Another averaging step is applied to get a single value for RWC across the entire layer. The resulting value corresponds to the magnitude of change in the layer's weights over training steps. Smaller changes over a prolonged period indicate that the layer's weights are nearing an optimum. 

\subsection{Analyzing RWC through Clustering and Dimensionality Reduction}
\label{clustering}
Most Deep Neural Networks include millions or even billions of parameters. Modern architectures stack hundreds of layers, making interpretability difficult for layerwise approaches like RWC. We use clustering to find similarities between the RWC trends between the layers of a network. For this paper, we are using K-Means Clustering \cite{lloyd:1982} with K++ algorithm \cite{kmeans++2007} to find layers with similar RWC learning trends. The input to the clustering algorithm is a matrix where layers are examples and their features are the RWC values per epoch (over training). These neighbors (layers) are then grouped into clusters based on their features/RWC values. Having reduced the 100+ layers into a mere 2-5 clusters makes investigating learning-dynamics in deep architectures feasible, and also partially automates the identification of similar behaving layer groups.  We can then compare these architectures with each other to understand how architectural differences and primitive choices can impact the learning-dynamics of a network. \\

\noindent Through our experiments we noticed that the certain RWC values, due to random-initialization and effects of backpropagation, had noisy outputs which created outliers. We remove these outliers by utilizing the mean and standard deviation of the sample. We then create a cut-off at 2 standard deviations surrounding the mean. This allows the clustering algorithm to perform better. This pre-processing stuff reduces the chances of clustering biases due to the presence of outlier without creating a big impact on the overall outcome of the number of clusters. 

\begin{equation} 
\label{Remove Outliers}
\centering  sample=\begin{cases}
			mean, & \text{if deviation > std}\\
            sample, & \text{otherwise}
		 \end{cases}
\end{equation}
Where $deviation = |sample - mean|$ and $mean$ \& $std$ refer to the mean and standard deviation of the sample.

In order to find the ideal number of clusters for K-Means, we used scree plots. Since our data is $n$ dimensional (where n is the number of epochs), we use Principal Component Analysis to reduce the data to 2-dimensions. Using this, we find the ideal number of clusters for K-Means using the RWC values and were able to effectively cluster the layers. These steps were repeated for each architecture-dataset pair presented in the paper.

\subsection{Experimental Settings}
\label{experimental_setting}

\textbf{Network Structure and Training} 
We use ResNet50 \cite{he2016deep} and the B4 version of the EfficientNet architecture \cite{tan2020efficientnet}. We chose these architectures for several reasons. ResNets are ubiquitously used in the research community for computer vision problems and EfficientNets (and its variants) hold the state-of-the-art performance on the ImageNet dataset \cite{russakovsky2015imagenet}. They also provide some variety in the information processing techniques and biases utilized to learn from images. For example, ResNets make use of residual connections, skip connections, and blockwise design while EfficientNets propose a novel model scaling method that utilizes a highly effective compound coefficient for scaling up CNNs in a more structured manner. Conventional approaches arbitrarily scale network dimensions, such as width, depth and resolution. EfficientNets uniformly scale each dimension with a fixed set of scaling coefficients. We incorporate these architectures for establishing some of the general trends we oberve in this work, and inconsistencies may be attributable to the concrete differences between them. We attempt to keep the computational complexity similar in terms of number of parameters. EfficientNet-B4 is significantly deeper as compared to ResNet-50 (in terms of number of layers). We utilize the opensource PyTorch \cite{PyTorch2019} implementation of EfficientNet for our experiments \cite{rw2019timm}

The general training strategies used for these architectures was kept consistent. We wished to keep the training settings similar for better comparison. We ran all the experiments on Google Colaboratory Pro. We utilize the Adam Optimizer with a learning rate of 1e-3 for all our experiments. All models are trained for 25 epochs. We utilize image sizes of 224x224 for EfficientNets as they are designed to work better with larger image sizes. ResNet-50 was trained with the original dataset sizes (32x32). We utilized a batch size of 32 (to accommodate for larger input sizes) for the EfficientNet experiments and 128 for the ResNet experiments.To interpret the layer-wise learning, we find the RWC as formulated in \ref{RWC} for each layer per epoch. We store the RWC array from each experiment, plot the the associated curves, and report the results in the following section.  

\textbf{Datasets} We use 3 benchmark datasets: CIFAR-10 \cite{Krizhevsky09learningmultiple} which contains 60,000 images of 10 classes, CIFAR-100 \cite{Krizhevsky09learningmultiple} which contains 60,000 images of 100 classes, and Street View House Numbers (SVHN) dataset \cite{netzer2011} which is similar to the MNIST datasets  \cite{lecun-mnisthandwrittendigit-2010}. As compared to MNIST, SVHN is significantly larger and poses a more difficult task for learning systems. These benchmark datasets see significant use in deep learning research. The datasets also provide good variety in the complexity of their associated learning tasks. CIFAR-10 and SVHN provide a decent level of complexity in terms of image content and task difficulty.CIFAR-100 has significantly more classes (compared to CIFAR-10) and fewer samples per class which makes it considerably more difficult. Our goal isn't to provide a state-of-the-art result on these datasets but rather to train a good model and understand the underlying learning trends.

\begin{table}
  \caption{Detailed hyperparameters used for training}
  \label{tab:training-hyperparameters}
  \centering
  \begin{tabular}{lllll}
    \toprule
    Architecture      & Datasets      & LR        & Image Size    & Batch Size \\
    \midrule
    ResNet50          & CIFAR-10      & $0.001$   & 32x32         & 128        \\
                      & CIFAR-100     & $0.001$   & 32x32         & 128        \\
                      & SVHN          & $0.001$   & 32x32         & 128        \\
    EfficientNet-B4  & CIFAR-10      & $0.001$   & 224x224       & 128        \\
                      & CIFAR-100     & $0.001$   & 224x224       & 128        \\
                      & SVHN          & $0.001$   & 224x224       & 128        \\
    \bottomrule
  \end{tabular}
\end{table}

\section{Results and Discussion}
\label{results}
\begin{table}[htb]
  \caption{Results}
  \label{tab:results}
  \centering
  \begin{tabular}{llllll}
    \toprule
    Architecture      & Datasets      & Train (Top-1)        & Train (Top-5)    & Test (Top-1)  & Test (Top-5) \\
    \midrule
    ResNet50          & CIFAR-10      & $96.1$               & $99.9$           & $90.0$        & $99.75$        \\
                      & CIFAR-100     & $92.4$               & $99.8$           & $62.4$        & $87.1$        \\
                      & SVHN          & $98.9$               & $99.9$           & $94.5$        & $99.3$        \\
    EfficientNet-B4  & CIFAR-10      & $97.9$               & $99.9$           & $89.1$        & $99.6$        \\
                      & CIFAR-100     & $95.1$               & $99.9$           & $62.3$        & $86.7$        \\
                      & SVHN          & $98.7$               & $99.9$           & $95.3$        & $99.4$        \\
    \bottomrule
  \end{tabular}
\end{table}

\subsection{ResNet-50}

\begin{figure}[hbt!]
\minipage{0.32\textwidth}
  \noindent\makebox[\textwidth]{\includegraphics[width=\linewidth]{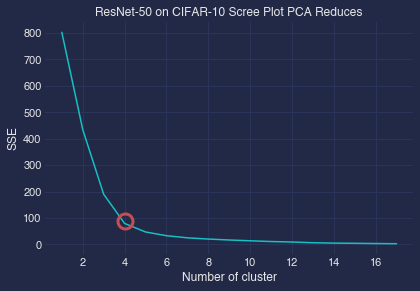}}
\endminipage\hfill
\minipage{0.32\textwidth}
  \noindent\makebox[\textwidth]{\includegraphics[width=\linewidth]{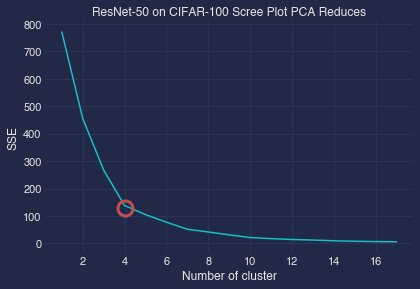}}
\endminipage\hfill
\minipage{0.32\textwidth}%
  \noindent\makebox[\textwidth]{\includegraphics[width=\linewidth]{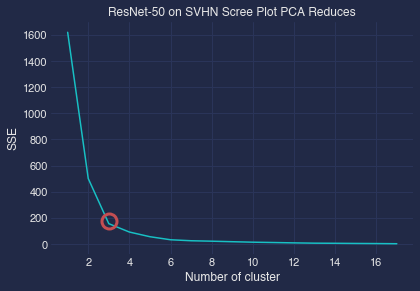}}
\endminipage
\caption{Optimal Clustering for the ResNet-50 model on CIFAR-10, CIFAR-100 and SVHN}
\label{fig:res-scree}
\end{figure}

\subsubsection{CIFAR-10}
\begin{figure}[hbt!]
    \centering
    \noindent\makebox[\textwidth]{\includegraphics[width=\linewidth]{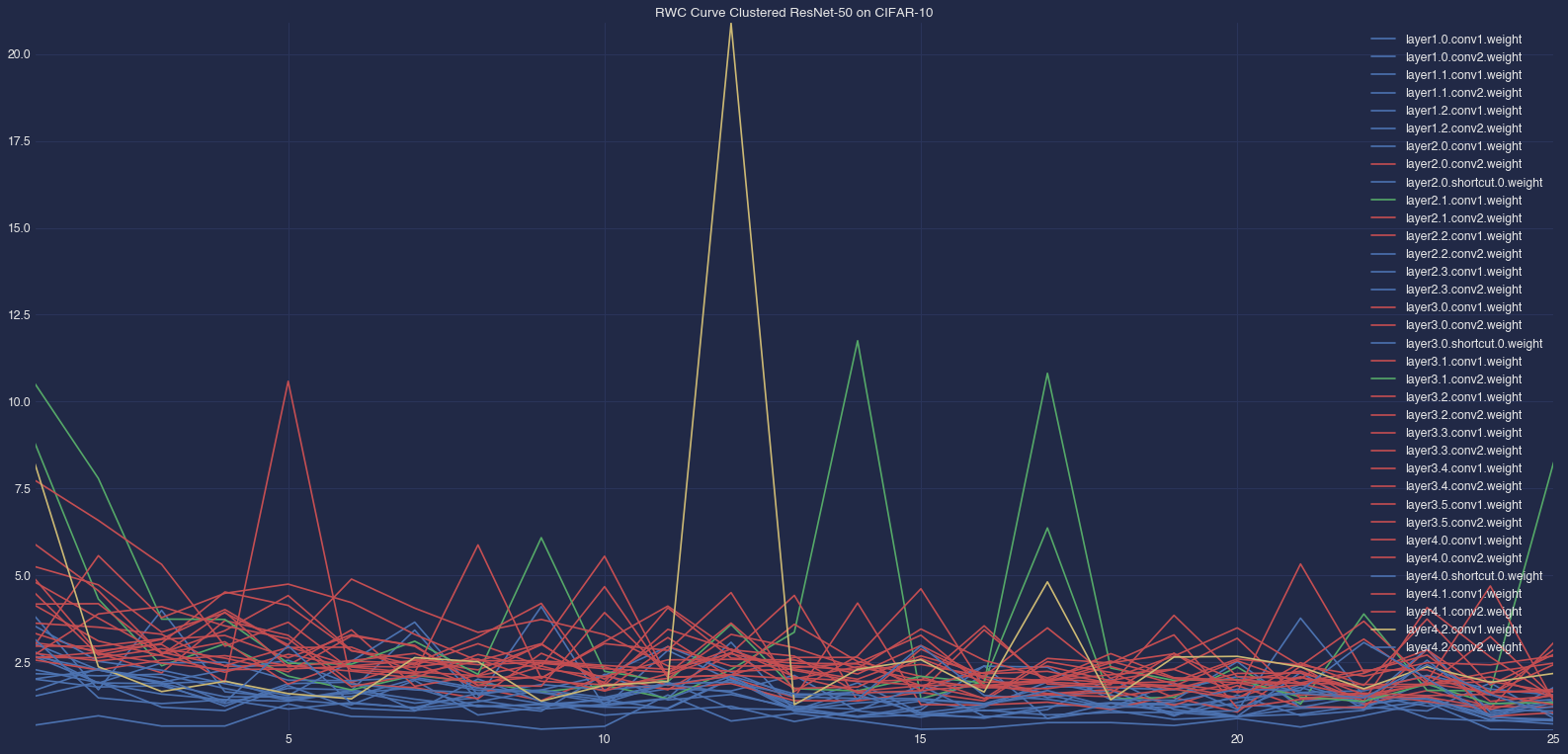}}
    \caption{Clustered RWC Curves for the ResNet-50 model on CIFAR-10}
    \label{fig:res-CIFAR-10}
\end{figure}

We notice that the model exhibits an decreased relative weight change in later layers of the network as compared to earlier layers Figure \ref{fig:res-CIFAR-10}. We notice that as we progress through the architecture layers over the training epochs the RWC decreases. There is a dip in the last few layers on average. This experiment achieved an accuracy of 90\% on the test set provided by the PyTorch distribution of CIFAR-10 \cite{torchvision}. We notice that the clustering approach is able to effectively separate layers based on their layer-wise learning dynamics. We think that the reason for low recruitment of later layers is due to the fact that the architecture is more than capable of handling a task like CIFAR-10. Task complexity might be the reason for this trend of high RWC in early and mid layers follower by lower RWC in the later layers. We also notice a cluster of layers (yellow) spike up more than others), this is an interesting trend due to the fact that the surround layers (belonging to other clusters) act differently. 

\subsubsection{CIFAR-100}
\begin{figure}[hbt!]
    \centering
    \noindent\makebox[\textwidth]{\includegraphics[width=\linewidth]{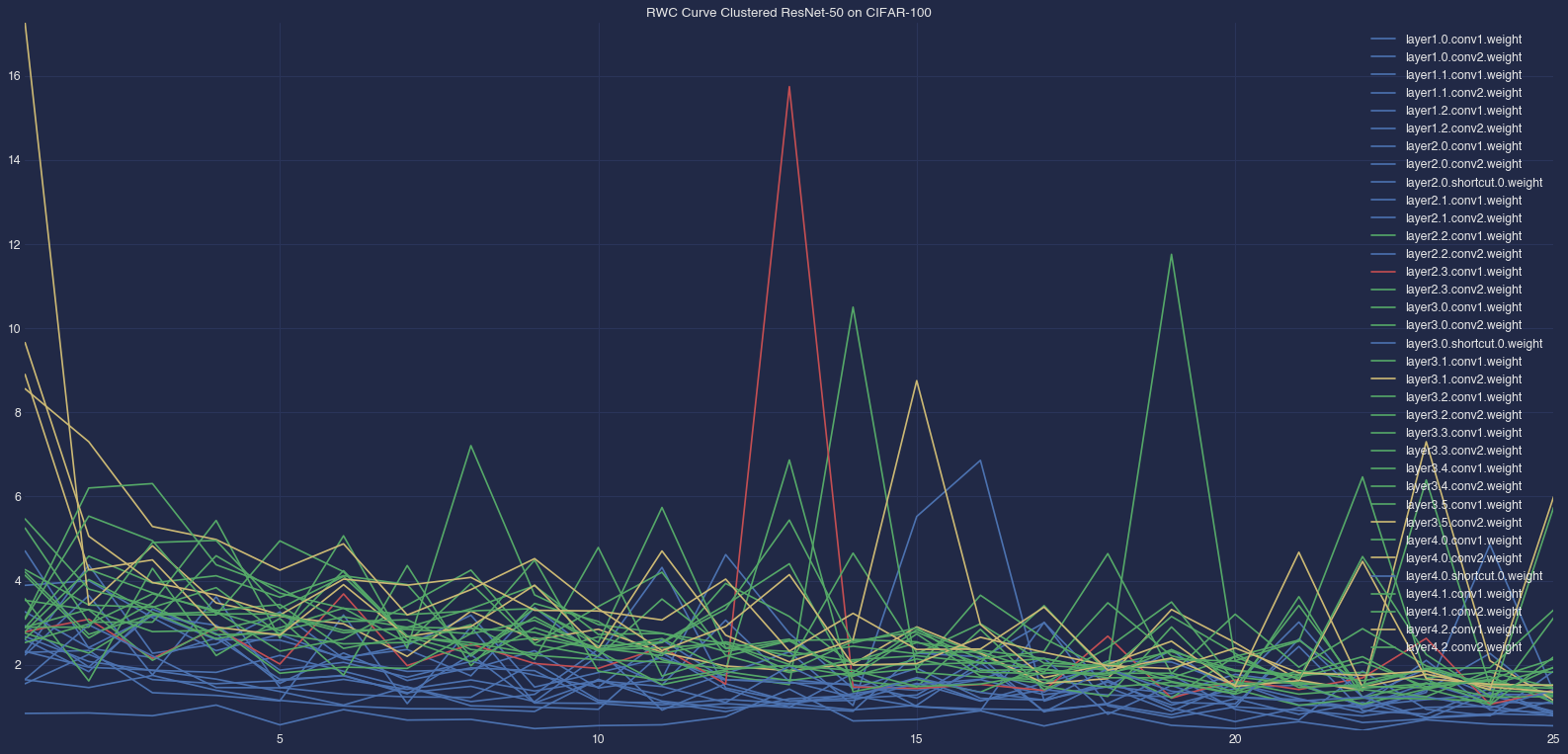}}
    \caption{Clustered RWC Curves for the ResNet-50 model on CIFAR-100}
    \label{fig:res-CIFAR-100}
\end{figure}
This dataset is more complex as compared to the CIFAR-10 variant and hence poses a bigger challenge for the network Figure \ref{fig:res-CIFAR-100}. This definitely makes a difference as we see a flip in the trends compared to CIFAR-10. We notice a sharp increase in RWC as we move up. This experiment achieved a test accuracy of 62.4\% on the test set provided by the PyTorch CIFAR-100 dataset \cite{torchvision}. The poor performance on this dataset suggests that the network is not able to learn a good representation for the task. This might be the reason that the later layers are recruited more as compared to CIFAR-10. Just like CIFAR-10 we notice an interesting cluster of layers (red). There is are massive RWC spikes during training which seem odd considering the curves for the surrounding layers. This might hint to two things, either this is random noise or that these layers might be doing something different compared to the others.

\subsubsection{SVHN}
\begin{figure}[hbt!]
    \centering
    \noindent\makebox[\textwidth]{\includegraphics[width=\linewidth]{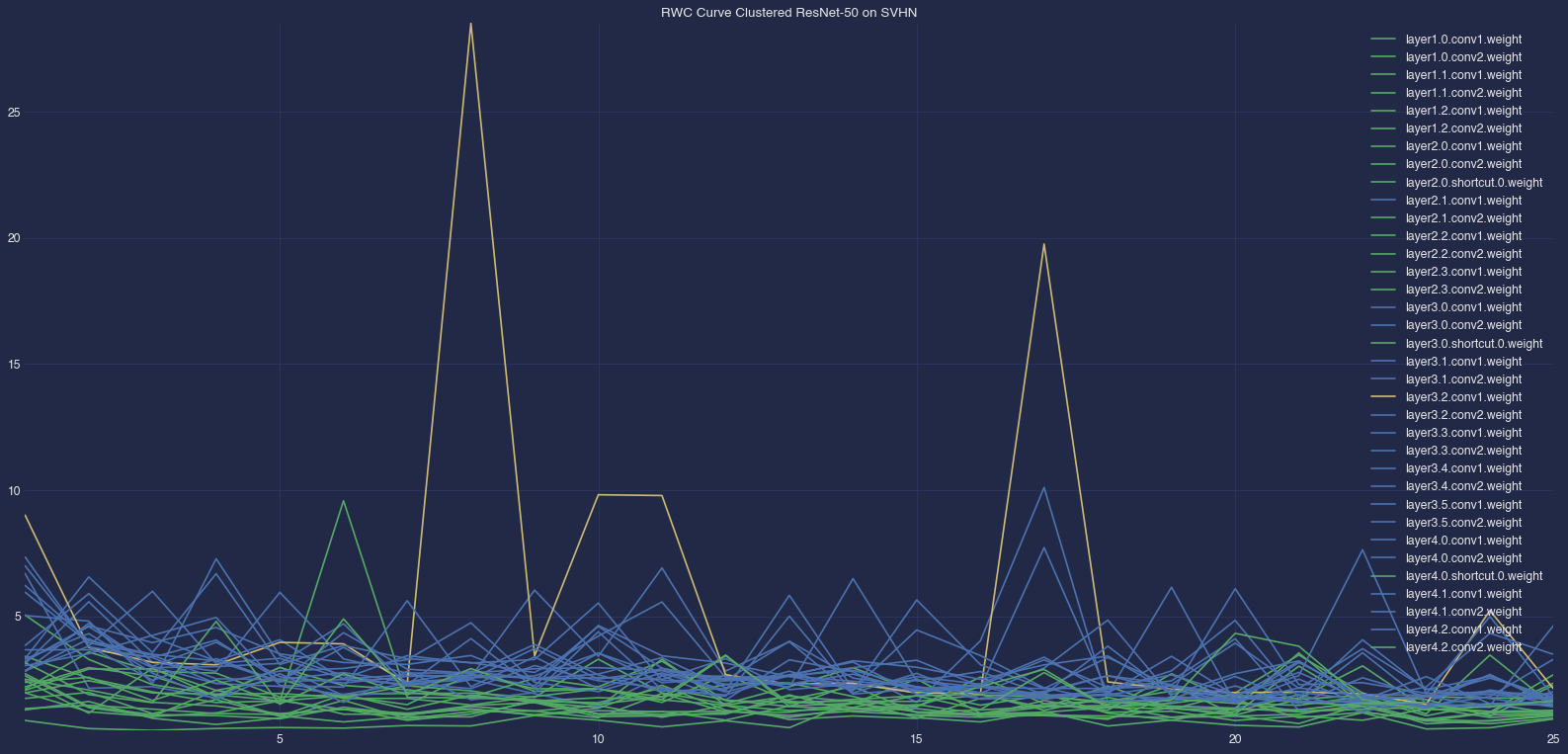}}
    \caption{Clustered RWC Curves for the ResNet-50 model on SVHN}
    \label{fig:res-svhn}
\end{figure}

SVHN is a simpler task and has a lot more data compared to the other two tasks. This makes it easier for the model to learn a good representation. This experiment manages to achieve a 94.5\% test accuracy. We can see in figure \ref{fig:res-svhn} trends from the clustering plots for this task are similar to CIFAR-10 in the sense that we notice a gradual increase as we move through the layers. One noticeable characteristic here is that the later layers are similar in terms of their change. Blocks 3 and 4 are a part of the same cluster due to their similar learning dynamic. There is an outlier cluster (yellow) that has massive spikes. This could just be random noise since it is only one layer.

\subsection{EfficientNet-B4}
EfficientNets utilize a special block based architectural primitive called MBConv Block \cite{sandler2019mobilenetv2}. This is essentially an inverted residual block (ResNets). Along with this inverted residual block the authors utilize depthwise convolutions for parameter efficiency. These perform the convolutional operation with a significantly smaller computational cost. EfficientNets also include Squeeze and Excitation to these blocks \cite{hu2019squeezeandexcitation}. Since the architecture is built using multiple different convolutional primitives we separate our analysis by splitting the layers by these primitives.

\begin{itemize}
    \item Depthwise Convolutional Layers
    \item Pointwise Convolutional Layers
    \item Squeeze and Excitation (expand and reduce convolutions)
    \item Pointwise Linear Expansion Layers (used with Squeeze and Excite)
\end{itemize}

By separating the layers in our analysis we can better compare the learning dynamics for the architecture.

\begin{figure}[hbt!]
\minipage{0.32\textwidth}
  \noindent\makebox[\textwidth]{\includegraphics[width=\linewidth]{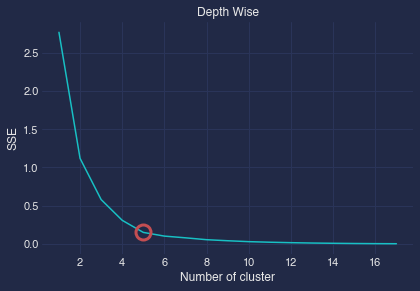}}
\endminipage\hfill
\minipage{0.32\textwidth}
  \noindent\makebox[\textwidth]{\includegraphics[width=\linewidth]{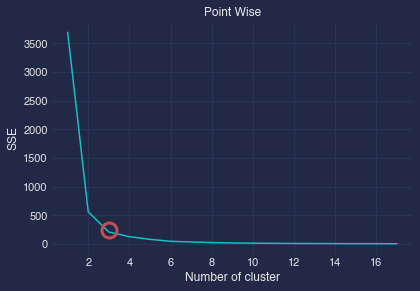}}
\endminipage\hfill
\minipage{0.32\textwidth}%
  \noindent\makebox[\textwidth]{\includegraphics[width=\linewidth]{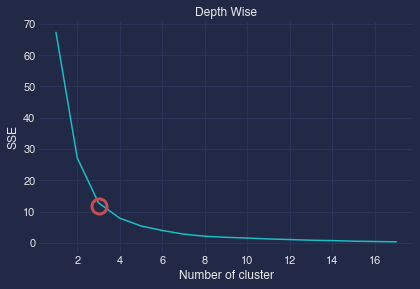}}
\endminipage
\caption{Optimal Clustering for for different Efficient-Net-B4 convolutional primitives on CIFAR-10, CIFAR-100 and SVHN}
\label{fig:eff-scree}
\end{figure}

\pagebreak 
\subsubsection{CIFAR-10}

\begin{figure}[hbt!]
    \centering
    \noindent\makebox[\textwidth]{\includegraphics[width=\linewidth]{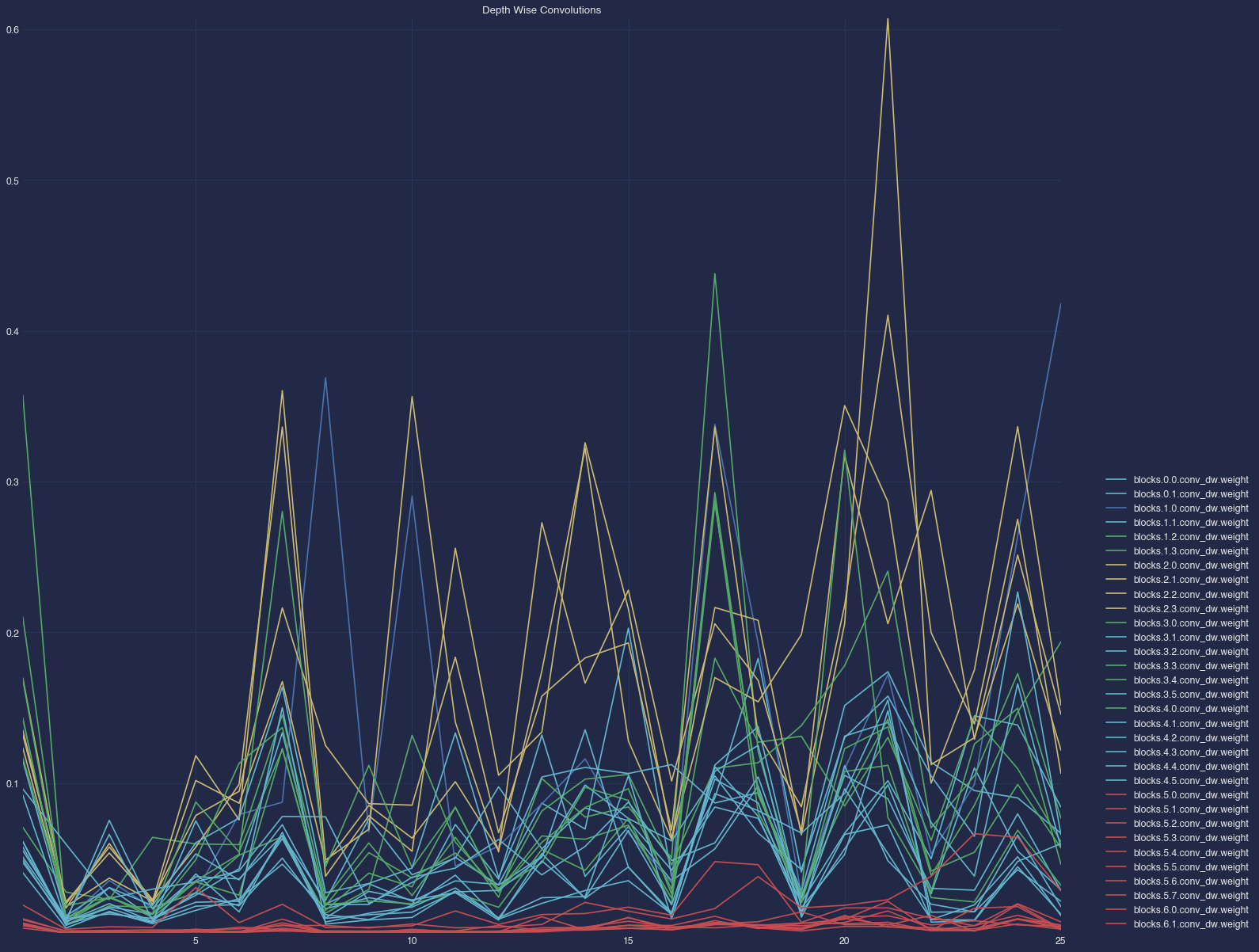}}
    \caption{Clustered RWC Curves for the Efficient-Net-B4 Depth-wise Convolutions on CIFAR-10}
    \label{fig:eff-CIFAR-10}
\end{figure}
As discussed earlier we split the EfficientNet model by its convolutional primitives. We observed trends for all of them across training. We choose to include the most interesting ones in our analysis. We can see in Figure \ref{fig:eff-CIFAR-10} that our clustering technique manages to group sets of layers that have similar relative weight change. We notices that the later layers have minimal changes for the CIFAR-10 task. We also notice that the earlier and middle layers have the most amount of changes. We believe that the task complexity might be the reason for this trend. Lower recruitment in later layers on easier classification tasks was observed in previous work as well \cite{agrawal2021investigating}. CIFAR-10 is a simple task and hence the architecture does not recruit the later layers as much. The performance is great and these trends corroborate this finding. There are a few layers that stand out as a separate cluster (dark blue lines). We call these layers outliers as they are surrounded by layers that learn similarly but these layers themselves have a different learning pattern. We believe that our framework allows us to detect these kind of special cases. 

\subsubsection{CIFAR-100}
\begin{figure}[hbt!]
    \centering
    \noindent\makebox[\textwidth]{\includegraphics[width=\linewidth]{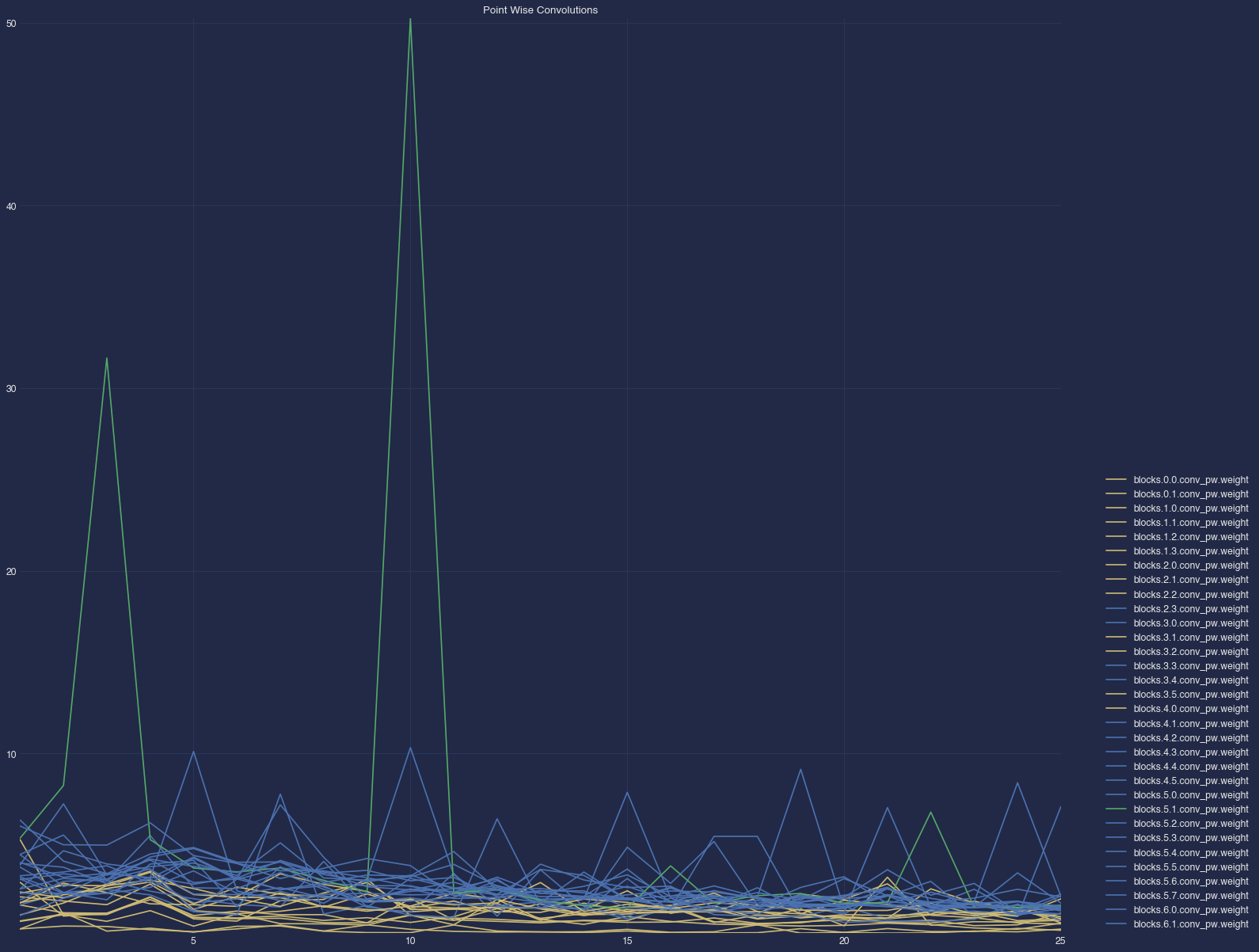}}
    \caption{Clustered RWC Curves for the Efficient-Net-B4 Point-wise Convolutions on CIFAR-100}
    \label{fig:eff-CIFAR-100}
\end{figure}

For the CIFAR-100 task we chose to discuss the point-wise convolutional primitive. We observe in Figure \ref{fig:eff-CIFAR-100} that the later layers are activated much more than the earlier layers. The clustering approach neatly bunches these layers together. This again points to the effect of task-complexity. CIFAR-100 is a complex task and this experiment does not achieve very high accuracy. Our curves support this point as we see a lot of recruitment in the later layers. Like CIFAR-10 we also notice some outlier layers which have massive spikes in the middle of training. 

\subsubsection{SVHN}

\begin{figure}[hbt!]
    \centering
    \noindent\makebox[\textwidth]{\includegraphics[width=\linewidth]{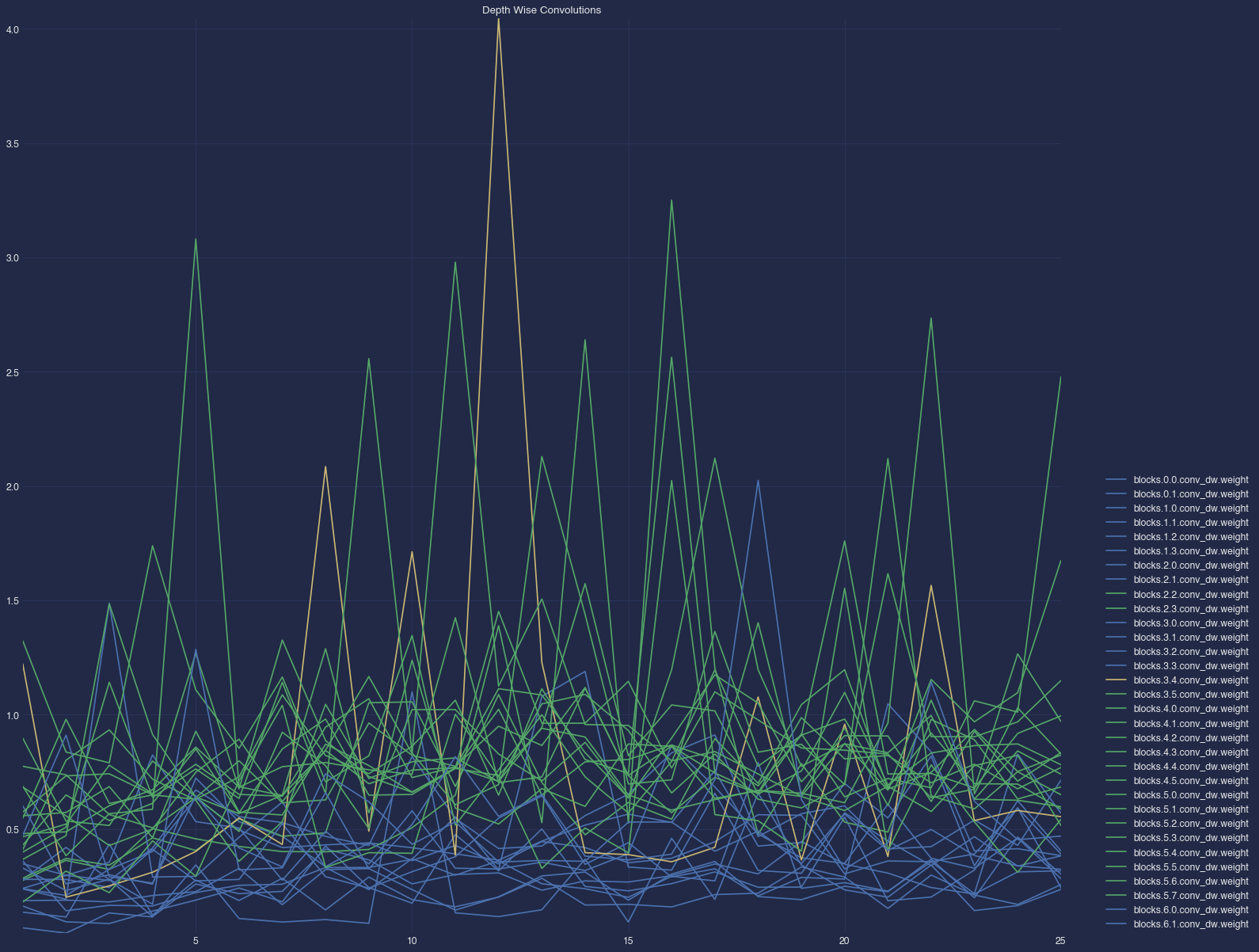}}
    \caption{Clustered RWC Curves for the Efficient-Net-B4 Depth-wise Convolutions on SVHN}
    \label{fig:eff-svhn}
\end{figure}

In terms of task complexity SVHN and CIFAR-10 are similar, infact one can consider SVHN to be an easier task due to its size and the type of classes. In this experiment we present the Depth-wise convolutional trend as it provides some interesting insights when coupled with the CIFAR-10 observation. Figure \ref{fig:eff-svhn} shows a similar trend to CIFAR-10 in the sense that the early to middle layers are activated the most. There is a slight difference in the case of SVHN where we notice that a lot more of the later layers are being utilized while the final couple of layers have a relatively small amount of change. We believe that this might hint towards a learning pattern observed in networks that get close to optimal accuracy.

\section{Limitations}
\label{limitations}
While our clustering approach using K-Means provided a boost to RWC interpretability metric, and also allowed us to better interpret much deeper and larger architectures, there are a couple limitations to our current work that open up room for further research and improvement. K-Means struggles when the data is very noisy and has outliers present. In an ideal scenario, the outliers should be completely ignored from clustering but K-Means fails to do so. To remedy this situation, we manually removed outliers from our RWC values which improved the performance of K-Means significantly. 

\noindent Clustering is an efficient and quick way to group the layers in a network based on their learning trends but we presume that other methods might exist that can allow for a much better visualization into the learning of a neural network. RWC as a metric, can be further improved to better distinguish the learning trends between layers as the current difference in RWC values between layers is quite small.

\newpage

\section{Conclusion and Future Work}
\label{conclusion}
In this paper we improve prior work done using the RWC metric as an interpretability tool by wrapping it with a clustering and dimensionality reduction approach that helps dynamic analysis of improve layer-wise learning. We apply our approach to deep and complex state-of-the-art architectures over various datasets. We report trends that are observed during these experiments and provide insights into their underlying mechanisms. 

This work forms a first step towards improving interpretability of neural networks on a layer level. Future directions for this work include analyzing learning trends in ImageNet scale datasets, new modalities like NLPs, and a variety of architectures. We hope that our work can provide another tool for researchers to investigate the inner workings of neural networks.

\newpage
\bibliography{refs}
\bibliographystyle{unsrt}






\end{document}